\documentclass[letterpaper, 10 pt, conference]{Packages/ieeeconf}  

\IEEEoverridecommandlockouts                              
 
\usepackage{float}
\usepackage{times}
\usepackage{epsfig}
\usepackage{graphicx}
\usepackage{amsmath}
\usepackage{amssymb}
\usepackage[ruled]{algorithm}
\usepackage{algpseudocode}
\usepackage{ctable}
\usepackage[export]{adjustbox}
\usepackage{import}
\usepackage{pgf}
\usepackage{setspace}
\usepackage{url}
\usepackage{hyperref}
\usepackage{capt-of,etoolbox}
\usepackage{pifont} 
\newcommand{\xmark}{\ding{55}} 

\makeatletter
\let\NAT@parse\undefined
\makeatother
\usepackage[numbers]{natbib}

\usepackage{Packages/sparo_acronyms}
\usepackage{Packages/sparo_math}
\usepackage{Packages/sparo_SIunits}
\usepackage{Packages/sparo_misc}
\usepackage{multirow}
\usepackage{indentfirst}

\usepackage{soul,color}
\usepackage{verbatim} 

\usepackage{float}

\definecolor{gl}{HTML}{008000}

\title{\LARGE \bf
   PoLaRIS Dataset:\\ A Maritime Object Detection and Tracking Dataset in Pohang Canal
}

\author{Jiwon Choi$^{1*}$, Dongjin Cho$^{1*}$, Gihyeon Lee$^{1}$, Hogyun Kim$^{1}$, Geonmo Yang$^{1}$, Joowan Kim$^{2}$, and Younggun Cho$^{1\dagger}$
	\thanks{$^{1*}$Jiwon Choi, $^{1*}$Dongjin Cho, $^{1}$Gihyeon Lee, $^{1}$Hogyun Kim, $^{1}$Geonmo Yang and $^{1\dagger}$Younggun Cho are with the Electrical and Computer Engineering, Inha University, Incheon, South Korea. {\tt\small [jiwon2, d22g66, leekh951, hg.kim, ygm7422]@inha.edu, yg.cho@inha.ac.kr} $^{2}$Joowan Kim is with the Samsung Heavy Industry, Daejeon, South Korea.{\tt\small joowani.kim@samsung.com} \hfill \break 
  (*) represents equal contribution. 
  }%
}

\begin{document}

\maketitle
\begin{abstract} 
Maritime environments often present hazardous situations due to factors such as moving ships or buoys, which become obstacles under the influence of waves. In such challenging conditions, the ability to detect and track potentially hazardous objects is critical for the safe navigation of marine robots. To address the scarcity of comprehensive datasets capturing these dynamic scenarios, we introduce a new multi-modal dataset that includes image and point-wise annotations of maritime hazards. Our dataset provides detailed ground truth for obstacle detection and tracking, including objects as small as 10$\times$10 pixels, which are crucial for maritime safety.
To validate the dataset's effectiveness as a reliable benchmark, we conducted evaluations using various methodologies, including \ac{SOTA} techniques for object detection and tracking. These evaluations are expected to contribute to performance improvements, particularly in the complex maritime environment. To the best of our knowledge, this is the first dataset offering multi-modal annotations specifically tailored to maritime environments.
Our dataset is available at \texttt{\url{https://sites.google.com/view/polaris-dataset}}.
\end{abstract}
\section{Introduction}


In maritime environments, \ac{USVs} require precise object recognition to ensure safe autonomous navigation. However, \ac{USVs} face diverse challenges such as irregular lighting conditions and unpredictable obstacles. 
Therefore, datasets that account for these factors are crucial for advancing autonomous maritime navigation systems. 
Although existing maritime datasets \cite{chung2023pohang, 9381638, liu2021efficient, nanda2022kolomverse, prasad2017video, Cheng_2021_ICCV, bovcon2019mastr1325} have been used as a foundation for \ac{USVs} navigation research, they suffer from two major limitations.


\begin{figure}[t]
    \centering
    \def\width{0.48\textwidth}%
        {%
     \includegraphics[clip, trim= 0 20 0 10, width=0.48\textwidth]{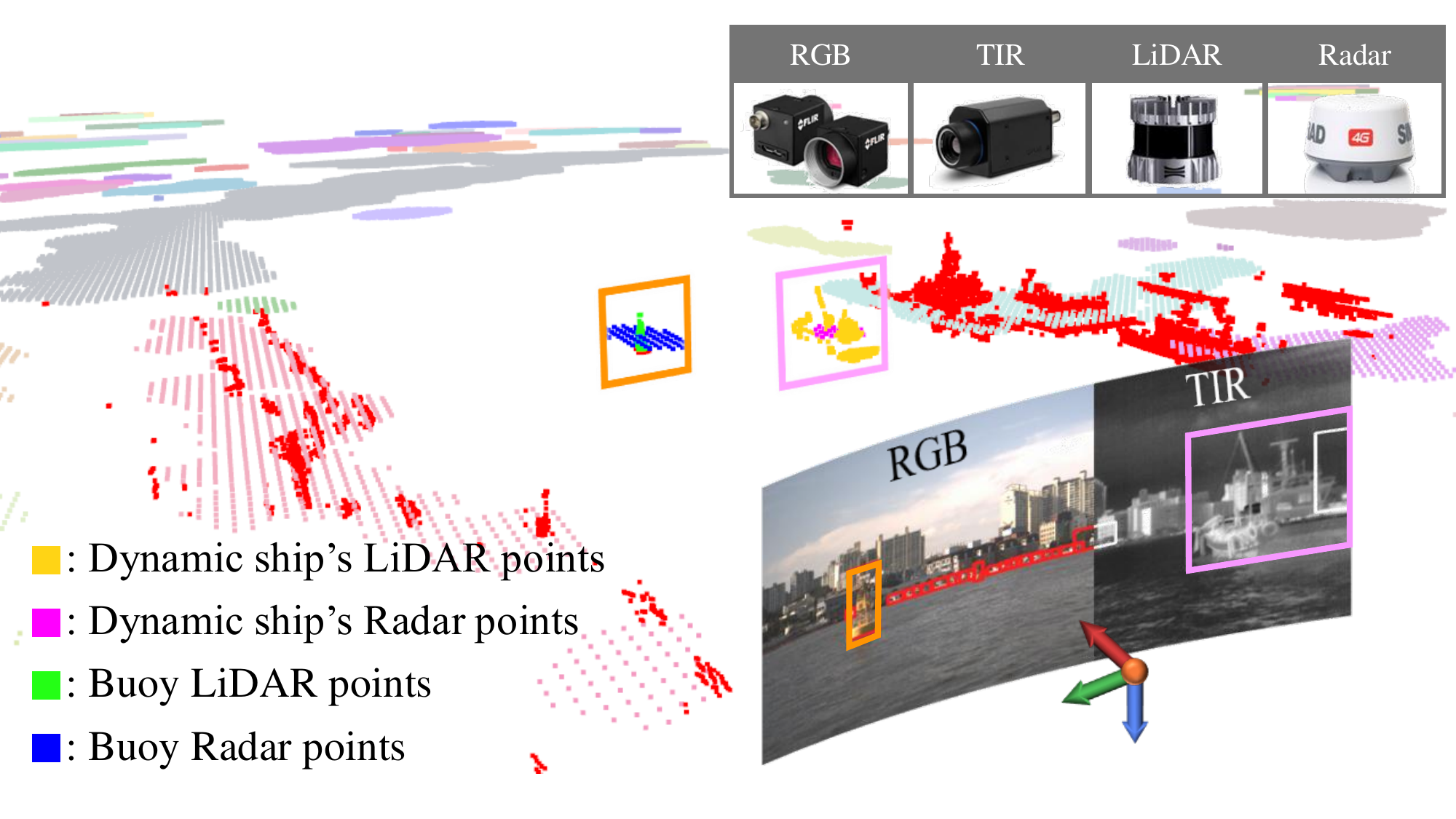}
        }
    \vspace{-0.55cm}
    \caption{\texttt{Pohang00} sequence's \texttt{11194}th scene. 
    A clustered sparse Radar point cloud is scattered far and wide in the background.
    A dense red LiDAR point cloud is also scattered relatively short.
    To convert this scene \texttt{PoLaRIS00}, we first annotate a RGB image.
    Then, annotated bounding boxes in the RGB image are mapped to TIR.
    Finally, we extract only the LiDAR and Radar point clouds projected on annotated bounding boxes of the RGB image to achieve multi-modal annotation.
    }
    \vspace{-0.55cm}
    \label{fig:main}
\end{figure}

First, \ac{USVs} need to accurately detect and track dynamic objects that can disrupt their navigation. However, under limited lighting conditions, such as in cases of light reflection or low visibility 
 \cite{almansoori2020analysis, zhang2015vais}, RGB cameras are less reliable, making it essential to utilize other sensors beyond the visible spectrum. In addition, precise localization of these obstacles is critical for effective collision avoidance. While multi-sensor data integration is key to addressing these challenges, few datasets comprehensively incorporate such elements. 
 

Second, it is crucial to detect dynamic objects early due to the long stopping distances of \ac{USVs}. However, existing datasets often lack accurate information on long-range or small objects, particularly in terms of tracking data, which limits their usefulness in real-world scenarios.

As shown in \figref{fig:main}, to address these limitations and support advancements in autonomous maritime navigation, we introduce a new dataset that provides multi-modal object annotations and ground truth for obstacle tracking in \ac{USVs} operations. The primary contributions of our work are as follows:

\begin{table*}[t]

\caption{Various Maritime Datasets}
\centering\resizebox{\textwidth}{!}{
{\tiny
\begin{tabular}{c||c|c|c|c|c|c|c|c|c}
\toprule
\hline
\multirow{2}{*}{\textbf{Datasets}} & \multicolumn{4}{c|}{\textbf{Bounding box}} & \multicolumn{3}{c|}{\textbf{Tracking Annotation}}  & \multirow{2}{*}{\textbf{Image size}} & \multirow{2}{*}{\textbf{Session}} \\ \cline{2-8}
           & \multicolumn{1}{c|}{Left}       & \multicolumn{1}{c|}{Right}      & \multicolumn{1}{c|}{TIR}        & Dynamic    & \multicolumn{1}{c|}{LiDAR} 
           & \multicolumn{1}{c|}{Radar}      & Image                           &                                 &    \\ \hline
USVInland \cite{9381638}              & \multicolumn{1}{c|}{\xmark}     & \multicolumn{1}{c|}{\xmark}     & \multicolumn{1}{c|}{\xmark}     & \xmark     & \multicolumn{1}{c|}{$\bigtriangleup$} 
                                      & \multicolumn{1}{c|}{\xmark}     & \xmark                          & 640 $\times$ 480                & Day  \\
MID \cite{liu2021efficient}           & \multicolumn{1}{c|}{\checkmark} & \multicolumn{1}{c|}{\xmark}     & \multicolumn{1}{c|}{\xmark}     & \checkmark & \multicolumn{1}{c|}{\xmark} 
                                      & \multicolumn{1}{c|}{\xmark}     & \xmark                          & 640  $\times$ 480               & Day \\
KOLOMVERSE \cite{nanda2022kolomverse} & \multicolumn{1}{c|}{\checkmark} & \multicolumn{1}{c|}{\xmark}     & \multicolumn{1}{c|}{\xmark}     & \xmark     & \multicolumn{1}{c|}{\xmark} 
                                      & \multicolumn{1}{c|}{\xmark}     & \xmark                          & 3840 $\times$ 2160              & Day/Night \\
FVessel \cite{guo2023asynchronous}    & \multicolumn{1}{c|}{\checkmark} & \multicolumn{1}{c|}{\xmark}     & \multicolumn{1}{c|}{\xmark}     & \checkmark & \multicolumn{1}{c|}{\xmark}                   
                                      & \multicolumn{1}{c|}{\xmark}     & \checkmark                      & 2560 $\times$ 1440              & Day                      \\
Singapore \cite{prasad2017video}      & \multicolumn{1}{c|}{\checkmark} & \multicolumn{1}{c|}{\xmark}     & \multicolumn{1}{c|}{\checkmark} & \checkmark & \multicolumn{1}{c|}{\xmark}                   
                                      & \multicolumn{1}{c|}{\xmark}     & \checkmark                      & 1920 $\times$ 1080              & Day                      \\
Flow \cite{Cheng_2021_ICCV}           & \multicolumn{1}{c|}{\checkmark} & \multicolumn{1}{c|}{\xmark}     & \multicolumn{1}{c|}{\xmark}     & \checkmark & \multicolumn{1}{c|}{\xmark}                   
                                      & \multicolumn{1}{c|}{\checkmark} & \xmark                          & 1280 $\times$ 720               & Day/Night                \\
MODD2 \cite{bovcon2019mastr1325}      & \multicolumn{1}{c|}{\checkmark} & \multicolumn{1}{c|}{\checkmark} & \multicolumn{1}{c|}{\xmark}     & \xmark     & \multicolumn{1}{c|}{\xmark}
                                      & \multicolumn{1}{c|}{\xmark}     & \xmark                          & 1278 $\times$ 958               & Day                      \\ \hline
                                      
\textbf{Ours}       & \multicolumn{1}{c|}{\checkmark} & \multicolumn{1}{c|}{\checkmark} & \multicolumn{1}{c|}{\checkmark} & \checkmark & \multicolumn{1}{c|}{\checkmark}
           & \multicolumn{1}{c|}{\checkmark} & \checkmark                      & \textbf{2048 $\times$ 1080}              & \textbf{Day/Night}                 \\
\hline
\bottomrule
\multicolumn{10}{r}{$\bigtriangleup$ means that data is provided but label is not provided.} \\
\end{tabular}}}
\label{tab:related}
\vspace{-0.6cm}
\end{table*}

\begin{itemize}
    \item \textbf{Multi-Scale Object Annotation}:
    We offer scale-aware annotations from large to small objects. To streamline this process, we employ an existing object detector to generate initial annotations for large-scale objects, which are then manually refined, along with manual annotation of smaller-scale objects.

    \item \textbf{Dynamic Object Annotation for Tracking}:
    Our dataset includes tracking data for dynamic objects that may interfere with USV navigation, facilitating more accurate experimentation and improving object detection and tracking performance.
    
    \item \textbf{Multi-Modal Objects Annotation}: 
    In addition to providing image annotations for \ac{TIR} data, we offer point-wise annotations for \ac{LiDAR} and \ac{Radar} data. These annotations, applicable across different sessions, serve as depth references. 
    To improve efficiency, we employ a semi-automatic labeling approach, which significantly reduces the manual annotation workload while ensuring reliable ground truth through human verification.
    
    
    

    \item \textbf{Benchmark}: 
    To validate the utility of our dataset as a benchmark, we perform evaluations using conventional and \ac{SOTA} methods for object detection and tracking. These evaluations confirm the practical applicability of our dataset and demonstrate its potential to drive performance improvements in autonomous navigation systems within complex maritime environments.
\end{itemize}

The rest of our paper is as follows:
Section \uppercase\expandafter{\romannumeral2} introduces related works. 
Section \uppercase\expandafter{\romannumeral3} represents an overview of our labeling process. 
Section \uppercase\expandafter{\romannumeral4} consists of various benchmarks of our datasets. 
Finally, Section \uppercase\expandafter{\romannumeral5} and Section \uppercase\expandafter{\romannumeral6} indicate a discussion and a conclusion, respectively.

\section{Related Works}
In recent decades, influential datasets and benchmarks have been offered.
For instance, \textit{KITTI} dataset \cite{geiger2012we} provided synchronized multi-modal perception sensors including LiDAR and motion sensor data.
\textit{HeLiPR} dataset \cite{jung2023helipr} supplied various types of LiDAR data (e.g. spinning, non-repetitive, and solid-state type).
To increase the utility of these impactful datasets, \textit{SemanticKITTI} dataset \cite{behley2019semantickitti} contributed to the robotics community by presenting segmented point cloud labeling of the \textit{KITTI} dataset \cite{geiger2012we} and \textit{HeLiMOS} dataset \cite{lim2024helimos} presented diverse LiDAR type's point cloud labeling of dynamic objects in the \textit{HeLiPR} dataset \cite{jung2023helipr}.

Although the number of datasets at maritime is relatively small compared to ground, maritime datasets also have been published as shown in \tabref{tab:related}.
\citet{9381638} released a \textit{USVland} dataset which offers various perception and motion sensors similar to \textit{KITTI}  dataset \cite{geiger2012we}.
This dataset allows the testing of algorithms such as \ac{SLAM} but did not provide the object-level labeling required for safe navigation. 
\textit{KOLOMVERSE} \cite{nanda2022kolomverse} and \textit{MID} \cite{liu2021efficient} included many labeled objects under various conditions such as season, weather, and waves.
However, it is not suitable for testing real-world navigation-level algorithms such as collision avoidance because it does not provide depth information like \ac{LiDAR} or \ac{Radar}. 
\citet{guo2023asynchronous} proposed a automatic identification system (AIS) dataset called \textit{FVessel}.
This dataset is also not suitable for testing real-world navigation-level algorithms, because it lacks the roll and pitch motions caused by ever-changing waves in the \ac{USVs}.
\textit{MODD2} dataset \cite{bovcon2019mastr1325} provided both left and right image labelings, and the \textit{Singapore Maritime} dataset \cite{prasad2017video} provided a thermal image labeling along with a RGB image labeling.
However, they also didn't provide the labeled object's depth information.
\citet{Cheng_2021_ICCV} provided a \textit{Flow} dataset containing a small object that is approximately 2.5$\%$ the size of the image. 
Although they didn't provide a 3D bounding box including \ac{Radar}, they proposed a fusion technique of \ac{Radar}-vision algorithm to establish a new paradigm for object detection in maritime.
However, the lack of a 3D bounding box makes it difficult for users to evaluate the detection algorithm, and the \ac{Radar} depth is significantly less reliable than \ac{LiDAR} due to speckle noise and multipath. 

To compensate these absences, we propose a \textit{PoLaRIS} presented a multi-modal 3D object's bounding box with multi-modal depth based on the \textit{Pohang Canal} dataset \cite{chung2023pohang} inspired by the contribution of \textit{SemanticKITTI} \cite{behley2019semantickitti} and \textit{HeLiMOS} datasets \cite{lim2024helimos}.
We also provide a dynamic object, tracking ground truth, and perform various benchmarks.

\section{POLARIS Dataset}

\begin{figure}[ht]
    \centering
    \def\width{0.48\textwidth}
        {
     \includegraphics[clip, trim= 0 0 0 0, width=0.48\textwidth]{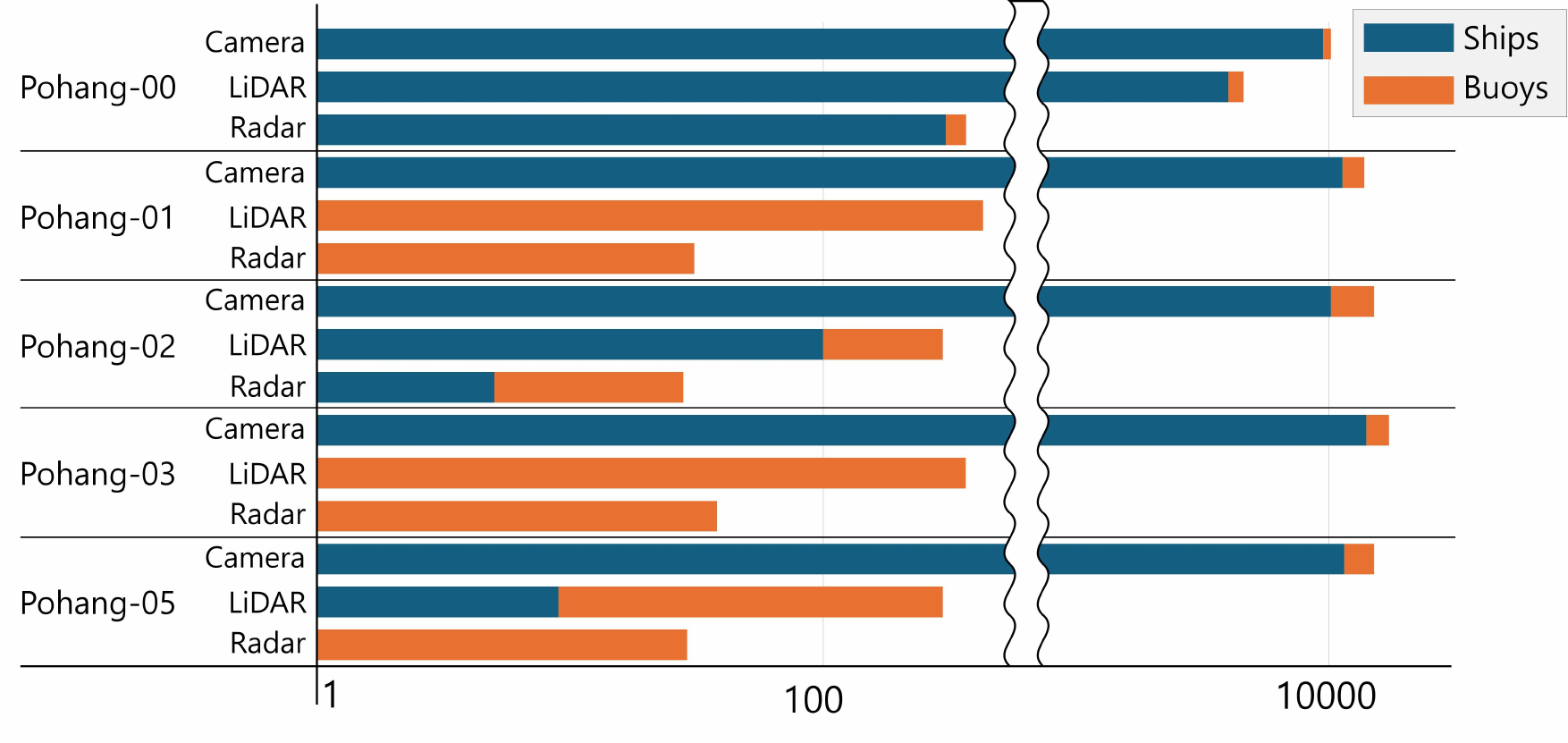}
        }
    \vspace{-0.5cm}
    \caption{The vertical axis represents the \texttt{Pohang00-04} sequences for the sensor modalities: image, \ac{LiDAR}, and \ac{Radar}. The horizontal axis indicates the number of labeled data for each sensor modality. \textit{Camera} refers to left, right, and \ac{TIR} image data, while \ac{Radar} and \ac{LiDAR} primarily detect dynamic obstacles and have limitations in identifying distant objects, resulting in significantly fewer data points compared to image data.}
     \vspace{-0.1cm}
    \label{fig:label_num}
\end{figure}

\begin{figure*}[t]
    \centering
    \def\width{\textwidth}
    {
        \includegraphics[width=\textwidth]{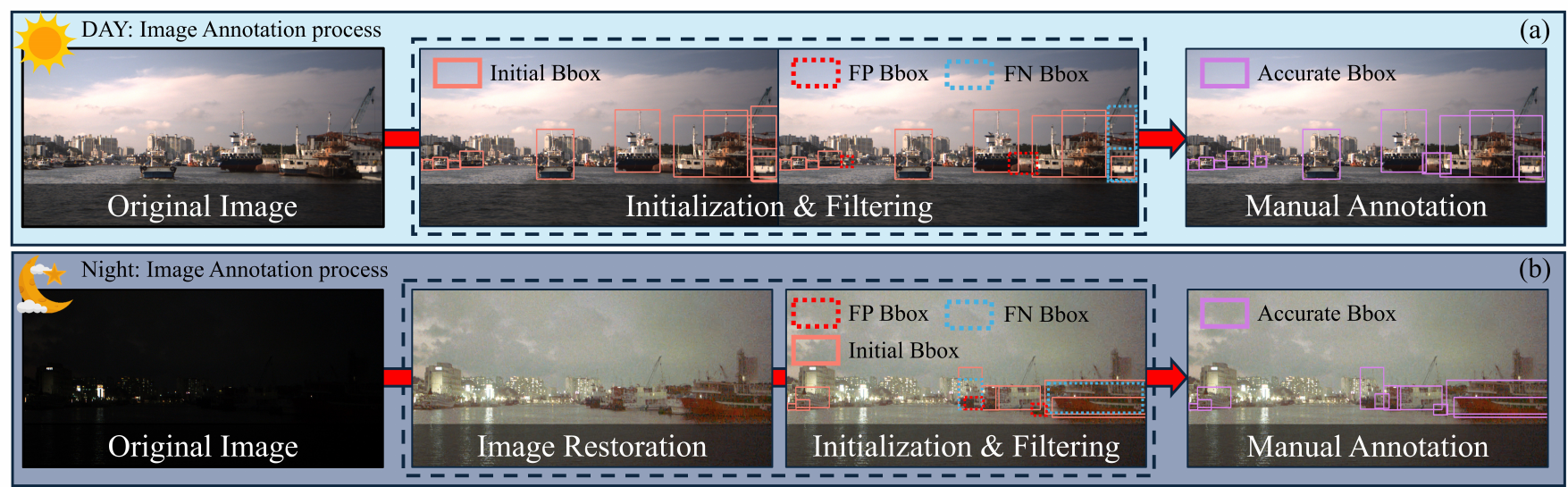}
    }
    \vspace{-0.55cm}
    \caption{The process of annotating the left image. (a) shows the annotation process during the day, where objects are more easily detected and labeled. In contrast, (b) illustrates the annotation process at night, requiring additional steps like image restoration due to low visibility. The Initial Bounding box (Bbox) results from the initialization step, while FP and FN \ac{Bbox} represent false positive and false negative detections, respectively. The Accurate \ac{Bbox} corresponds to the provided ground truth labels.}
    \vspace{-0.50cm}
    \label{fig:main_1}
\end{figure*}

Our dataset is based on the \textit{Pohang Canal} dataset \cite{chung2023pohang}, which consists of 6 sequences. We selected five major sequences \texttt{Pohang00,02,03,04} for day and \texttt{Pohang01} for night. \texttt{Pohang05} was excluded due to the absence of dynamic objects and poor lighting conditions. These five sequences encompass approximately 360,000 images (e.g., stereo images and TIR) with around 190,000 labels.

Our dataset provides multi-modal information, as shown in \figref{fig:label_num}, including both static and dynamic object annotations. It also includes ground truth for small-scale objects, as small as 10$\times$10 pixels (0.0045$\%$ of the image size). In addition, we provide tracking ground truth for obstacles, along with depth reference data extracted from ranging sensors, such as \ac{LiDAR} and \ac{Radar}.

To streamline the annotation process, we first annotate accurate labels for the left images and use these as references for annotating data from the other sensors.

\subsection{Initial Data Annotation}

\subsubsection{Left RGB Images}
To ensure precise annotations for the left image of the stereo camera, we perform an initialization process and manual annotation, as shown in \figref{fig:main_1}.
By providing sample bounding boxes via the initialization process, the annotator's manual labor is exponentially reduced.
In this process, we manually sample one image for every 20 images from the entire image set to create a meaningful training set. The remaining images are then used to form the test set, with sample bounding boxes generated using YOLOv8 \cite{yolov8_ultralytics}.
Then, a filtering process is applied. Using an \ac{IoU}-based method, calculated as
\begin{equation}
\text{IoU}(A, B) = \frac{|A \cap B|}{|A \cup B|},
\end{equation}
where $ A $ and $ B $ are initial bounding boxes from the same image, and overlapping boxes are removed if $ \text{IoU}(A, B) \geq 0.8 $.
After the initialization and filtering processes, we manually annotate all small objects that are less than 5$\%$ of the total image size (e.g. 2048 $\times$ 1080) in all images.

For \textbf{Day} sequence, we assign all the labels for the left image using the initialization process and manual annotation, as shown in \figref{fig:main_1}-(a).

For \textbf{Night} sequence, due to insufficient lighting conditions, it is difficult to distinguish objects in images.
To address this limitation, some researchers attempted to use \ac{TIR} cameras to detect objects \cite{krivsto2020thermal}.
However, annotating precise object labels remains challenging due to the inherent blurring and low resolution of \ac{TIR} camera outputs. 
To address this limitation, we utilize restored RGB images for the annotation process, rather than relying on TIR images. 
For image restoration, we employ a diffusion-based GSAD approach \cite{hou2024global}, which represents the \ac{SOTA} in low-light image enhancement.
After that, we manually annotate the restored images, as shown in \figref{fig:main_1}-(b).

\subsubsection{Right RGB Images}
We utilize the ground truth labels from the left image to extract the initial labels of the entire sequences. 
Since there is a difference in the \ac{FOV} between the left and right cameras, manual annotation is performed to generate ground truth labels for the right camera.

\begin{figure*}[ht]
    \centering
    \def\width{\textwidth}
    {
        \includegraphics[width=\textwidth]{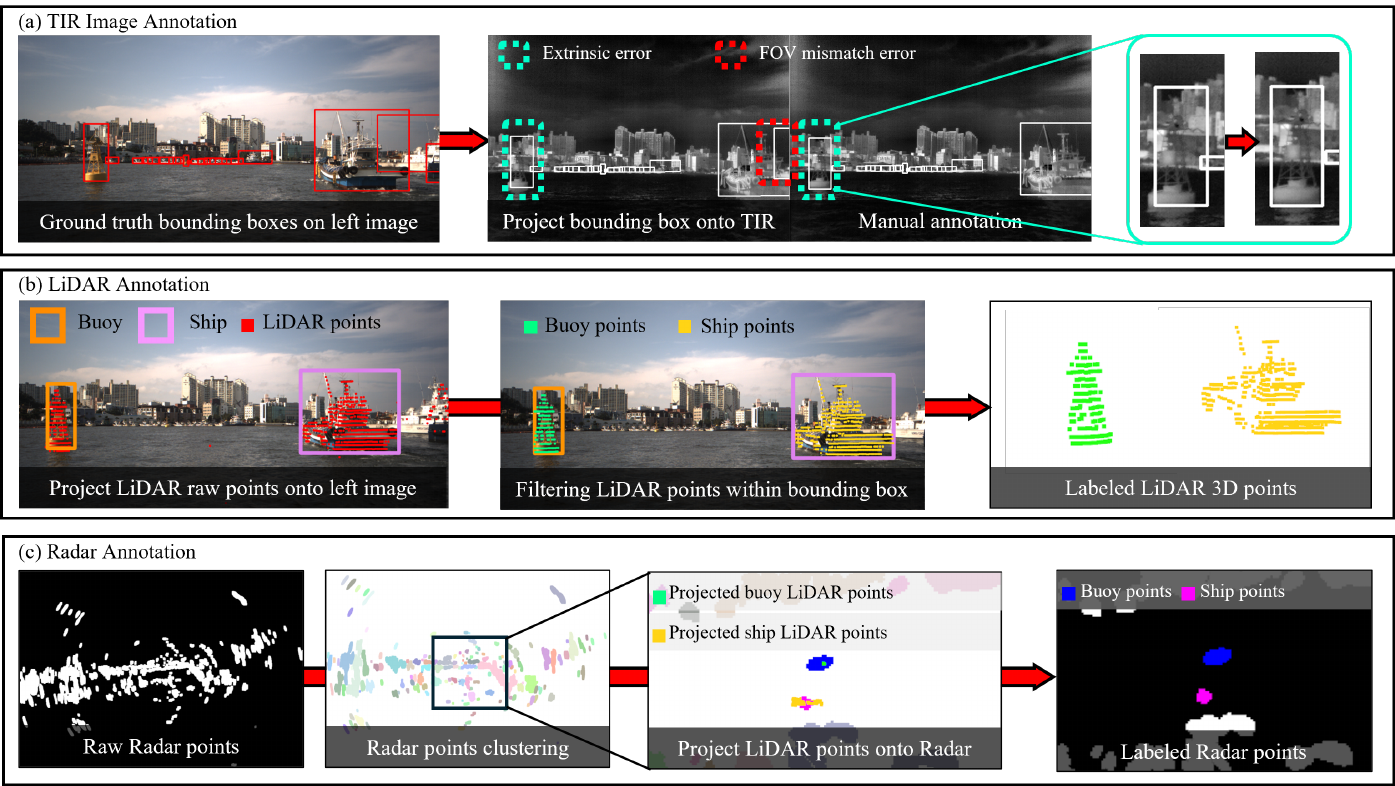}
    }
    \vspace{-0.50cm}
    \caption{The process of semi-automatic annotation for multi-modal sensors.
             (a) shows the process of defining labels in the TIR image using transformation data from the left image and manually correcting label errors.
             (b) shows the process of filtering LiDAR points based on the ground truth labels from the left image to obtain LiDAR points' annotation.
             (c) demonstrates the process of clustering Radar points and defining their labels by identify clusters that overlap with the labeled points obtained in (b).}

    \vspace{-0.50cm}
    \label{fig:semi_auto}
\end{figure*}

\subsection{Semi-Automatic Annotation}

\subsubsection{TIR Images} 
We leverage extrinsic parameters between TIR camera and left camera for efficient annotation. 
With the ground truth labels from the left images, it is feasible to obtain the annotation of the TIR camera using the transformation between the two cameras, as shown in the TIR Image Annotation section of \figref{fig:semi_auto}-(a).

To briefly explain the two transformations, the bounding box corner in the TIR image is followed as: 
\begin{equation}
 bb_t = K_t \cdot T_l^t \cdot K_l^{-1} \cdot bb_l,
\end{equation}
where $T_l^t$ is the transformation between the TIR and left cameras,
$bb$ represents the bounding box corner
(e.g. $\begin{bmatrix} u & v & 1 \end{bmatrix}^T$), and $K$ is the camera's intrinsic matrix, with subscripts $t$ and $l$ indicating the TIR and left cameras

The converted bounding box is used as the reference label for \ac{TIR} image.
However, due to transformation errors and \ac{FOV} differences between sensors, post-processing on the reference label is necessary.
For this purpose, we need to distinguish the object, but TIR images are inevitably a 16-bit.
Therefore, to address this, we used Fieldscale \cite{gil2024fieldscale} to convert the 16-bit TIR image to 8-bit for improved visualization, and then applied manual annotations to ensure accurate labeling.
This process is demonstrated in the manual annotations section shown in the TIR Image Annotation section of \figref{fig:semi_auto}-(a).

\subsubsection{LiDAR Points} 
We generate point-wise annotations for \ac{LiDAR} sensor data on obstacles, based on the accurate label information from the left image.
First, we project the \ac{LiDAR} points onto the left image based on the transformation between the left camera and \ac{LiDAR}.
Then, we remove all projected \ac{LiDAR} points outside the bounding box in the left image, as demonstrated in the \ac{LiDAR} Annotation section of \figref{fig:semi_auto}-(b).
However, the rectangular bounding box does not perfectly match the object's shape, some non-object areas may be included.
As a result, defining point labels based on the bounding box can lead to false positives.
To enhance accuracy, we manually verify and remove points within the bounding box that do not belong to the object. 
This process ensures that the 3D point labels more accurately represent the object's shape and position.

\subsubsection{Radar Points} 
It is difficult to directly fuse the camera and \ac{Radar} to generate annotations because the camera captures 2D forward-view images, while \ac{Radar} collects data in a 2D \ac{BEV} format.
To overcome this limitation, we fuse the \ac{Radar} with \ac{LiDAR}, which provides 3D information, instead of directly using the camera. 
Then, we generate point-wise \ac{Radar} annotations for dynamic objects, based on the existing \ac{LiDAR} annotations as shown in the Radar Annotation section of \figref{fig:semi_auto}-(c). 
The process for point-wise annotation of \ac{Radar} consists of the following steps:

\begin{enumerate}
    \item Convert the labeled \ac{LiDAR} points to \ac{Radar} coordinates, and project them in \ac{BEV} format.
    \item Identify the overlap region between the projected annotation \ac{LiDAR} points and the \ac{Radar} data in 3D space.
    Due to the wide horizontal beamwidth of the marine radar sensor \cite{bole2013radar}, it is not possible to assign point-wise annotations to areas that overlap with the \ac{LiDAR} points.
    \item Apply the DBSCAN \cite{ester1996density} algorithm to cluster the Radar data, and assign point-wise annotation to the \ac{Radar} clusters that overlap with the labeled \ac{LiDAR} points.
\end{enumerate}

This process reflects the characteristics of the \ac{Radar} sensor and contributes to improving annotation accuracy.

\subsection{Dynamic Object Tracking}
In our dataset, dynamic objects (e.g. moving ships and buoys) are defined as obstacles.
Then, they are assigned unique identifiers (IDs) during the tracking process.

We aim to provide tracking ground truth for multi-scale objects.
\begin{figure}[ht]
    \centering
    \def\width{0.48\textwidth}
        {
     \includegraphics[clip, trim= 70 185 65 190, width=0.48\textwidth]{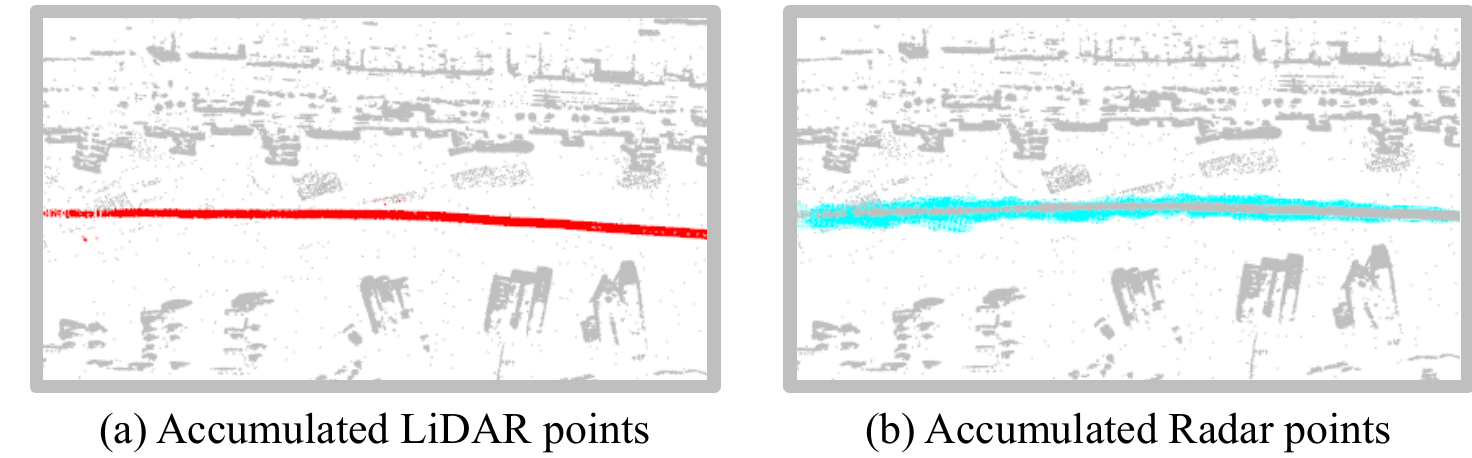}
        }
    \vspace{-0.6cm}
    \caption{
    Illustration of dynamic objects.
    }
        \vspace{-0.3cm}
    \label{fig:dyn_map}
\end{figure}
Therefore, we generate tracking ground truth from the moment of the stage of object detection, even for dynamic objects as small as 10$\times$10 pixels. 
By providing tracking ground truth data for small, distant, and obscured objects, our dataset can validate experiments on objects of various sizes.
To verify this, several evaluations are conducted in Section \uppercase\expandafter{\romannumeral4}.

In addition, by combining 2D tracking ground truth with the 3D labeled points from both LiDAR and Radar, we can provide precise metrics as shown in \figref{fig:dyn_map}. 

\subsection{File Structure}
 
\begin{figure}[h]
    \centering
    \def\width{0.48\textwidth}
        {
     \includegraphics[clip, trim= 5 150 0 140, width=0.48\textwidth]{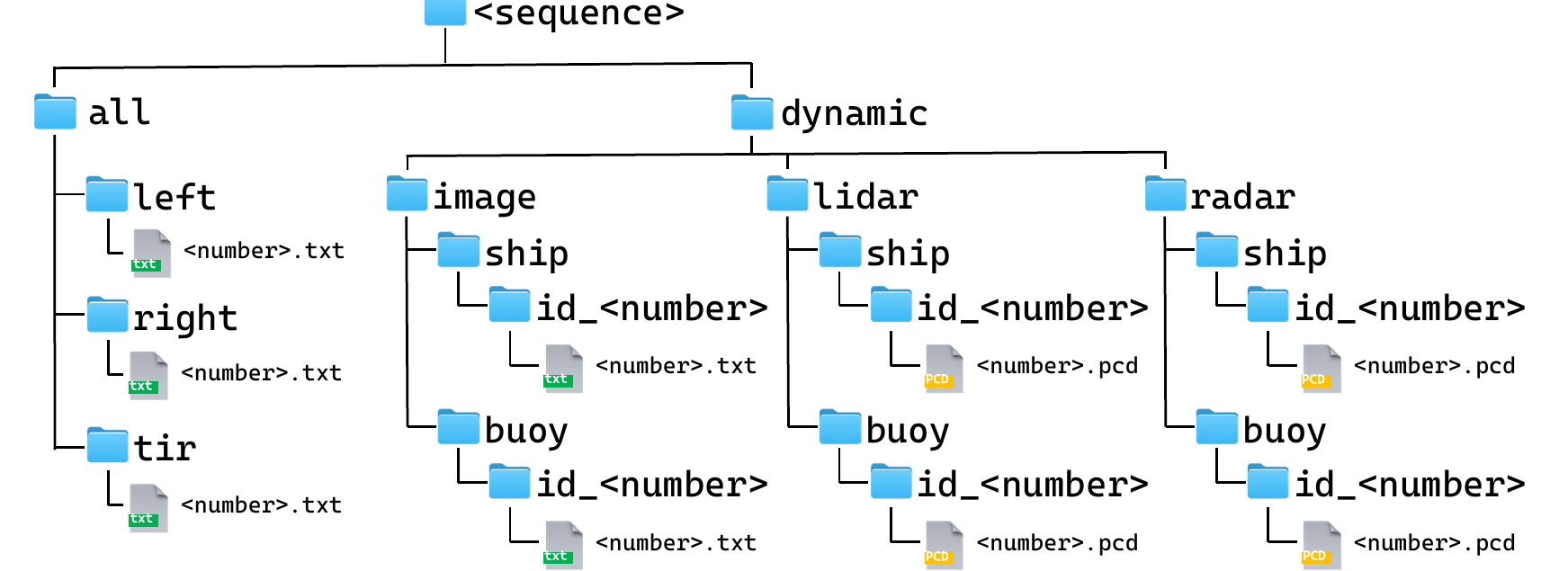}
        }
    \vspace{-0.5cm}
    \caption{File structure of our dataset.}
    \label{fig:file_format}
\end{figure}

All label files for the image sensors are stored in a folder named \textit{all}.  
The image labels and point-wise labels of dynamic objects are stored in the \textit{dynamic} folder, as shown in \figref{fig:file_format}.
\section{Benchmarks}

\begin{figure}[ht]
    \centering
    \def\width{0.5\textwidth}%
        {%
     \includegraphics[clip, trim= 0 0 0 0, width=0.48\textwidth]{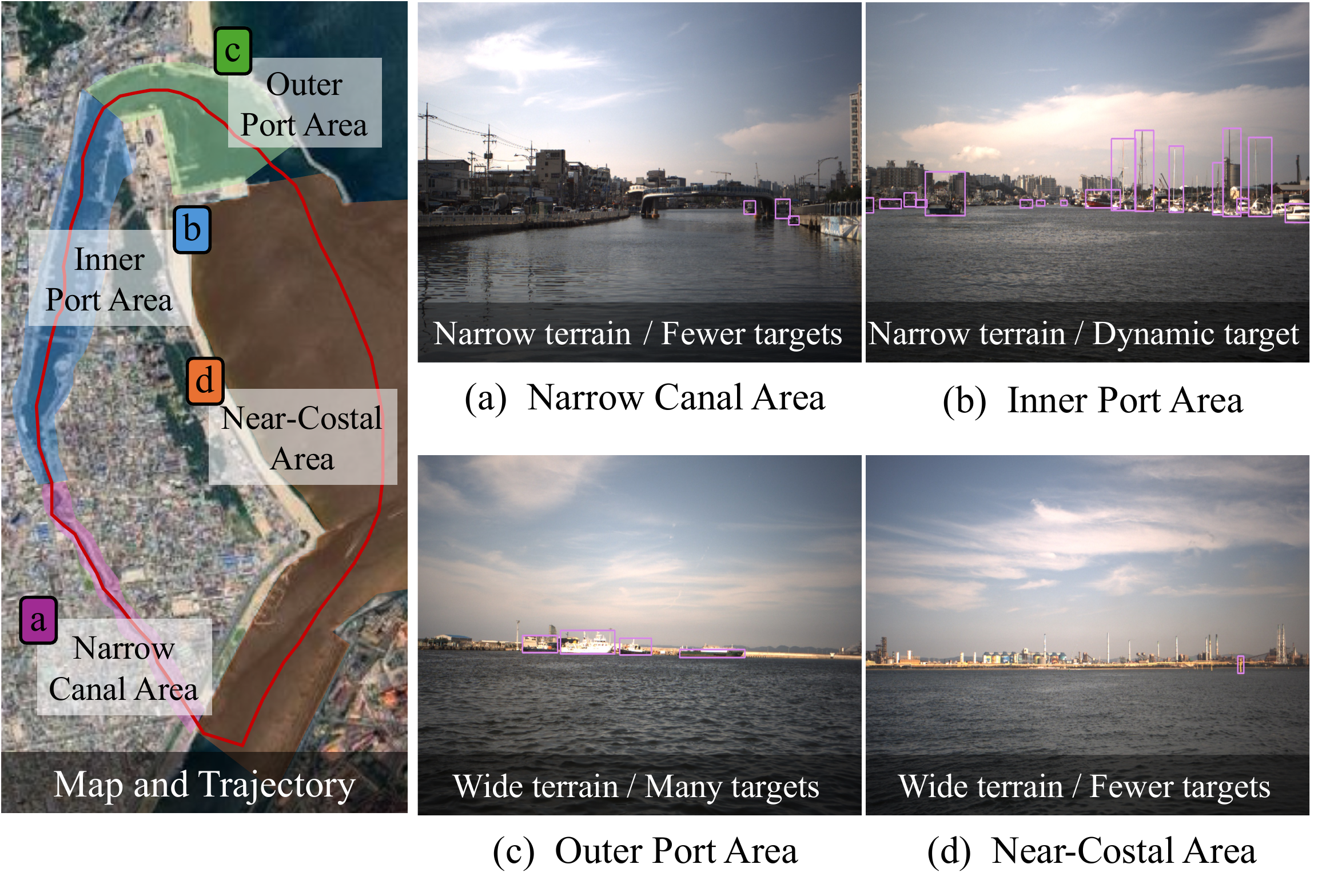}
        }
    \vspace{-0.62cm}
    \caption{The Pohang dataset \cite{chung2023pohang} trajectory is shown, divided into four regions: Narrow Canal Area, Inner Port Area, Outer Port Area, and Near-Coastal Area. Regions (b) and (c) contain more objects compared to Regions (a) and (d).
    }
    \vspace{-4mm}
    \label{fig:traj}
\end{figure}

To demonstrate validity for object detection and tracking across various modalities and lighting conditions, we conducted tests in the \texttt{Pohang00} sequence, which contains the highest number of multi-dynamic objects. 
For evaluation in low-light conditions and scenarios involving single-dynamic objects, we utilized the \texttt{Pohang01} sequence. 
Several metrics were used to assess performance on sequences containing obstacles. 
All scenarios utilized RGB and TIR images for detection and tracking evaluations, with TIR images converted to 8-bit using \ac{SOTA} methods from Fieldscale \cite{gil2024fieldscale}.
For detection evaluations, we sampled 5$\%$ of images from each of Regions (a)-(d) in \figref{fig:traj} for training and the rest for test sets. All detection results were fixed, and we evaluated the performance of the trackers using YOLOv8 \cite{yolov8_ultralytics}.

\subsection{Evaluation Metrics}

\subsubsection{mAP}
The performance of the object detection model was evaluated using the \ac{mAP}, which is calculated based on precision and recall:
\begin{equation}
   \text{Precision} = \frac{\text{TP}}{\text{TP} + \text{FP}}, \quad \text{Recall} = \frac{\text{TP}}{\text{TP} + \text{FN}}, 
\end{equation}
\begin{equation}
   AP = \int_0^1 p(r) \, dr,
\end{equation}
where TP, FP, and FN represent true positives, false positives, and false negatives. 
The Average Precision (\ac{AP}) is defined as the area under the precision-recall curve, where $ p(r) $ denotes the precision at a given recall $ r $. 
\begin{equation}
   \text{mAP} = \frac{1}{N} \sum_{i=1}^{N} AP_i,
\end{equation}
The \ac{mAP} is then computed as the mean of the AP values across all classes, where $ N $ is the total number of classes, and $ AP_i $ is the Average Precision for class $ i $.

\subsubsection{MOTA}
The Multiple Object Tracking Accuracy \ac{MOTA} \cite{bernardin2008evaluating} is a standard metric used to evaluate the performance of tracking algorithms. It combines three types of errors: FP, FN, and identity switches (IDSW). \ac{MOTA} is defined as follows:
\begin{equation}
   MOTA = 1 - \frac{|FN| + |FP| + |IDSW|}{|gtDet|},
\end{equation}
where $ |FN| $ represents the number of missed detections, $ |FP| $ represents the number of incorrect detections, and $ |IDSW| $ is the number of identity switches. $ |gtDet| $ is the total number of ground truth detections. This metric penalizes tracking errors, with a higher \ac{MOTA} indicating better tracking performance.

\subsubsection{IDF1}
The IDF1 score \cite{ristani2016performance} measures the accuracy of the predicted trajectories in terms of identity preservation. It evaluates how well the tracker assigns the correct identity to each object throughout the video sequence. The IDF1 score is defined as:
\begin{equation}
   IDF1 = \frac{2 \times IDTP}{2 \times IDTP + IDFP + IDFN},
\end{equation}
where $ IDTP $, $ IDFP $, and $ IDFN $ represent identity true positives, false positives, and false negatives, respectively. The IDF1 metric focuses on identity-based accuracy, providing insights into the effectiveness of the tracking system in maintaining object identities over time.

Additionally, \ac{IDP} and \ac{IDR} are used to evaluate identity assignment performance:
\begin{equation}
   IDP = \frac{IDTP}{IDTP + IDFP}
   , IDR = \frac{IDTP}{IDTP + IDFN}
\end{equation}
IDP measures the precision of identity assignments, while IDR measures the recall of the identity assignments.


\begin{table*}[ht]
\caption{Evaluation - object detection}
\centering\resizebox{\textwidth}{!}{
{\tiny
\begin{tabular}{c|c||c|c|c|c|c|c||c|c|c|c|c|c}
\toprule
\hline
       \multicolumn{2}{c||}{} &\multicolumn{3}{c|}{\textbf{\texttt{Pohang00} (Day-RGB)}} & \multicolumn{3}{c||}{\textbf{\texttt{Pohang01} (Night-RGB)}} & \multicolumn{3}{c|}{\textbf{\texttt{Pohang00} (Day-TIR)}} & \multicolumn{3}{c}{\textbf{\texttt{Pohang01} (Night-TIR)}} \\ \hline
Method      & Pre-train          & mAP$\uparrow$ ($\%$)         & Ship         & Buoy        & mAP$\uparrow$ ($\%$)   & Ship         & Buoy        & mAP$\uparrow$ ($\%$)   & Ship         & Buoy        & mAP$\uparrow$ ($\%$)  & Ship        & Buoy      \\ \hline
YOLOv8-L    & {\xmark}              & 78.0        & 89.3         & 66.6        & 77.2       & 88.4     & 66.0       & \textbf{58.4}        &  \textbf{70.1}      & \textbf{46.7}     & \textbf{64.0} &  \textbf{71.6} & \textbf{56.5} \\
YOLOv8-L    & COCO                  & 79.6        & \textbf{91.0}  & 68.1        & \textbf{84.2 } & \textbf{91.3}     & \textbf{77.1} &  -          &   -        &  -      & - &  - & - \\
YOLOv10-L   & {\xmark}              & 75.6        & 87.6         & 63.5        & 72.4       & 85.9     &  58.9      & 54.9        &  66.3      & 43.6     & 59.1&  69.0 & 49.3   \\
 
YOLOv10-L    & COCO                 & \textbf{80.1} & 90.6         &  69.6       & 82.9       & 90.9     & 74.9       & -         &  -       & -       & - & - &  - \\
RT-DETR-R50  & {\xmark}             & 78.4        & 86.7         & \textbf{70.0}   & 69.1       & 77.6     & 60.7       & 53.4        & 60.9       & 45.9     &  55.2 &  62.1 & 48.3  \\ \hline
YOLO-World-L w/o training & O+G     & 1.5         & 2.8          & 0.2         & 0.7        & 1.1      &  0.0       & -         & -        & -       & - & - &  - \\
Grounding Dino w/o training & O+G+C & 2.3         & 4.3          & 0.3         & 0.7        & 1.4      &  0.0       & -         & -        & -       &  - & - &  -  \\ \hline
\bottomrule
\multicolumn{14}{r}{‘O’, ‘G’, and ‘C’ denote pertaining using Objects365 \cite{shao2019objects365}, GoldG \cite{hudson2019gqa, plummer2015flickr30k}, and Cap4M \cite{liu2023grounding} dataset.} 
\end{tabular}
\label{table:detection}}}
\vspace{-0.4cm}
\end{table*}

\subsection{Experimental Settings}
All models were trained using the AdamW \cite{loshchilov2017decoupled} optimizer (learning rate: 0.000714, momentum: 0.9) on an NVIDIA GeForce RTX 3090 GPU. From the \texttt{Pohang00} (during the day) and \texttt{Pohang01} (during the night) sequences, 2,158 and 2,399 images were sampled for training, and 959 and 1,066 images for validation, respectively.

\subsection{Object Detection}

\subsubsection{Implementation Details}
We evaluated object detection models using \ac{SOTA} methods from both the \ac{YOLO} and \ac{DETR} series. Specifically, we employed YOLOv8 \cite{yolov8_ultralytics}, YOLOv10 \cite{THU-MIGyolov10}, and RT-DETR \cite{zhao2024detrs} as closed-set models, alongside open-set models like YOLO-World \cite{cheng2024yolow} and Grounding DINO \cite{liu2023grounding} to assess the generalization capabilities of our dataset. For YOLOv8 \cite{yolov8_ultralytics} and YOLOv10 \cite{THU-MIGyolov10}, RGB images were resized to 736$\times$736, and TIR images to 512$\times$512. Due to the modality differences between TIR and RGB images, we did not use models pre-trained on the COCO \cite{lin2014microsoft} dataset for TIR experiments.

\subsubsection{Detection Results} 
\tabref{table:detection} summarizes the performance of the models trained on RGB and TIR images under varying lighting conditions, reporting \ac{mAP} scores over \ac{IoU} thresholds ranging from 0.50 to 0.95. 
In RGB-based detection, the models exhibited strong performance in the Ship class, with significant improvements in the nighttime Buoy class after COCO pre-training. 
However, the performance in TIR-based detection was comparatively lower, likely because the model was optimized for RGB images.
The open vocabulary models exhibited generally lower performance, which can be attributed to the challenging nature of the dataset, containing a large number of small objects.


\subsection{Object Tracking}

\begin{table}[ht]
\centering
\caption{Evaluation - object tracking}
\resizebox{0.48\textwidth}{!}{  
\begin{tabular}{c|c|c||ccccc}
\toprule
\hline

\textbf{Data}       & \textbf{sensor}      & \textbf{Method} &\multicolumn{1}{c|}{\textbf{MOTA$\uparrow$}} & \multicolumn{1}{c|}{\textbf{IDR$\uparrow$}} & 
\multicolumn{1}{c|}{\textbf{IDP$\uparrow$}} & \textbf{IDF1$\uparrow$} \\ \hline
\multirow{4}{*}{\texttt{Pohang00}} & \multirow{4}{*}{RGB} 
                                           & SORT & 90.9$\%$ & \textbf{99.9$\%$} & 91.1$\%$ & 95.3$\%$ \\
                    &                      & ByteTrack     & 96.0$\%$ & 99.7$\%$ & 96.3$\%$ & 98.0$\%$ \\
                    &                      & OC-SORT       & \textbf{98.5$\%$} & 99.8$\%$ & \textbf{98.6$\%$} & \textbf{99.2$\%$} \\
                    &                      & Hybrid-SORT   & 97.8$\%$ & 99.8$\%$ & 97.9$\%$ & 98.9$\%$ \\ \hline
\multirow{4}{*}{\texttt{Pohang01}} & \multirow{4}{*}{RGB}  
                                           & SORT          & 85.0$\%$ & 99.8$\%$ & 85.1$\%$ & 91.9$\%$ \\
                    &                      & ByteTrack     & \textbf{99.6$\%$} & 99.8$\%$ & \textbf{99.8$\%$} & \textbf{99.8$\%$} \\
                    &                      & OC-SORT       & 99.3$\%$ & \textbf{99.9$\%$} & 99.4$\%$ & 99.7$\%$ \\
                    &                      & Hybrid-SORT   & 98.2$\%$ & \textbf{99.9$\%$} &  98.2$\%$ & 99.1$\%$ \\ \hline
\multirow{4}{*}{\texttt{Pohang00}} & \multirow{4}{*}{TIR} 
                                           & SORT           & 62.5$\%$ &  \textbf{97.9$\%$} & 63.9$\%$ & 77.4$\%$ \\
                    &                      & ByteTrack      & 71.9$\%$ & 97.7$\%$  & 73.7$\%$ & 84.0$\%$ \\
                    &                      & OC-SORT        & \textbf{75.5$\%$} & 97.7$\%$  & \textbf{77.4$\%$} & \textbf{86.4$\%$} \\
                    &                      & Hybrid-SORT    & 74.8$\%$ & 97.7$\%$  & 76.6$\%$ & 85.9$\%$ \\ \hline
\multirow{4}{*}{\texttt{Pohang01}} & \multirow{4}{*}{TIR} 
                                           & SORT           & 71.3$\%$ & 94.0$\%$ & 76.1$\%$ & 84.1$\%$ \\
                    &                      & ByteTrack      & \textbf{88.0$\%$} & \textbf{94.0$\%$} & \textbf{94.0$\%$} & \textbf{94.0$\%$} \\
                    &                      & OC-SORT        & 82.5$\%$ & 93.1$\%$ & 89.2$\%$ & 91.1$\%$ \\
                    &                      & Hybrid-SORT    & 81.6$\%$ & 93.0$\%$ & 88.2$\%$ & 90.6$\%$ \\ \hline
\toprule
\hline
\end{tabular}
\label{tab:tracking}
\vspace{-0.5cm}
}
\end{table}
\subsubsection{Implementation Details}
We evaluated the performance of the tracker on dynamic objects using heuristic-based multi object tracking methods, including SORT \cite{bewley2016simple}, ByteTrack \cite{zhang2022bytetrack}, OC-SORT \cite{cao2023observation}, and Hybrid-SORT \cite{yang2024hybrid}.
All tracker experiments, YOLOv8-L (as presented in \tabref{table:detection}) was used as the detection model, with the \ac{IoU} threshold fixed at 0.3. To evaluate performance, we used the MOTA and IDF1 metrics to assess accuracy and ID matching.

\subsubsection{Tracking Results}
\tabref{tab:tracking} validates the object tracking methodology using RGB and TIR sensors in varying illumination conditions. 
While RGB-based trackers perform well day and night, TIR-based trackers underperform compared to RGB.
In addition, OC-SORT tracker performed better in the multi-dynamic objects sequence(\texttt{Pohang00}), while ByteTrack performed better in the single-dynamic objects sequence(\texttt{Pohang01}). 
These results confirm that the dataset is valid for multi-sensor object tracking tasks.
\section{Conclusion \& Discussion}
In this paper, we introduced \texttt{POLARIS}, a multi-modal dataset designed for object labeling in maritime environments. The dataset includes approximately 360,000 labeled images from RGB and TIR cameras, along with point-wise annotations from Radar and LiDAR. Tracking ground truths for dynamic objects are also provided, projected within bounding boxes, enabling the evaluation of detection and tracking algorithms.

One of the key challenges in maritime environments is the variability in object sizes and sensor ranges. \texttt{POLARIS} addresses this by offering detailed multi-modal annotations to improve navigation and object detection in dynamic maritime settings.
In future work, we will extend the dataset to cover more diverse environments and introduce semantic-level annotations alongside bounding box-level annotations to enhance its utility further.

\section{Acknowledgements}
We appreciate Prof. Jinhwan Kim's \textit{MORIN} group, particularly Dongha Chung, for publishing the \textit{Pohang Canal} dataset~\cite{chung2023pohang}.

\scriptsize
\bibliographystyle{Packages/IEEEtranN} 
\bibliography{Packages/string-short, Packages/references}

\end{document}